\documentclass{article}
\usepackage{ijcai22}

\usepackage{times}
\usepackage{soul}
\usepackage{url}
\usepackage[hidelinks]{hyperref}
\usepackage[utf8]{inputenc}
\usepackage[small]{caption}
\usepackage{graphicx}
\usepackage{amsmath}
\usepackage{amsthm}
\usepackage{booktabs}
\usepackage{algorithm}
\usepackage{algorithmic}
\usepackage{subfig}
\usepackage{enumerate}
\usepackage{latexsym}
\usepackage{graphicx}
\usepackage{times}
\usepackage{array}
\usepackage{comment}
\usepackage{multirow}
\usepackage{epstopdf}
\usepackage{url}
\usepackage{amsmath}
\usepackage{amssymb}
\usepackage{mathrsfs}
\usepackage{booktabs}

\newcommand\ourmethod{\textsc{TransLog}}
\title{\ourmethod{}: A Unified Transformer-based Framework for \\ Log Anomaly Detection}

\author{
}

\author{
Hongcheng Guo$^{1}$\thanks{Equal contribution.}\and
Xingyu Lin$^{4}$\textsuperscript{$\star$} \and
Jian Yang$^{1}$\and
Yi Zhuang$^{2}$\and
Jiaqi Bai$^{1}$\and
Tieqiao Zheng$^{3}$ \and
Liangfan Zheng$^{3}$ \and
Weichao Hou$^{3}$ \and
Bo Zhang$^{3}$ \and
Zhoujun Li$^{1}$\thanks{Corresponding author.} 
\affiliations
$^1$State Key Lab of Software Development Environment, Beihang University\\
$^2$Bio-robot and human-computer interaction laboratory, Waseda University\\
$^3$Cloudwise Research
$^4$University of Southern California
\emails
\{hongchengguo, jiaya, jiaqi, lizj\}@buaa.edu.cn,
linxingy@usc.edu,\\
syouyi2020@asagi.waseda.jp
\{steven.zheng,liangfan.zheng,william.hou,bowen.zhang\}@cloudwise.com
}
\begin{document}

\maketitle

\begin{abstract}
%不超过200个词
Log anomaly detection is a key component in the field of artificial intelligence for IT operations (AIOps). Considering log data of variant domains, retraining the whole network for unknown domains is inefficient in real industrial scenarios especially for low-resource domains. However, previous deep models merely focused on extracting the semantics of log sequences in the same domain, leading to poor generalization on multi-domain logs. To alleviate this issue, we propose a unified \textbf{Trans}former-based framework for \textbf{Log} anomaly detection (\ourmethod{}), which is comprised of the pretraining and adapter-based tuning stage. Our model is first pretrained on the source domain to obtain shared semantic knowledge of log data. Then, we transfer the pretrained model to the target domain via adapter-based tuning. The proposed method is evaluated on three public datasets including one source domain and two target domains. Experimental results demonstrate that our simple yet efficient approach, with fewer trainable parameters and lower training costs in the target domain, achieves state-of-the-art performance on three benchmarks\footnote{We will release the pretrained model and code.}. 

%避免再训练引起的时间和效率损耗

\end{abstract}

\section{Introduction}

%%%%%%%%%%%%%%%%%%%%%%%%%%Edited YJ%%%%%%%%%%%%%%%%%
With the rapid development of large-scale IT systems, numerous companies have an increasing demand for high-quality cloud services. Anomaly detection \cite{breier2015anomaly} is a critical substage to monitoring data peculiarly for logs, which describe detailed system events at runtime and the intention of users in the large-scale services \cite{zhang2015rapid}. The field of artificial intelligence for IT operations (AIOps) \cite{dang2019aiops} intends to empower IT operations by integrating advanced deep learning algorithms to meet these challenges, which ensures the stability of company data and maintain high efficiency simultaneously.

\begin{figure}[t]
    \centering {
        \includegraphics[width=1\columnwidth]{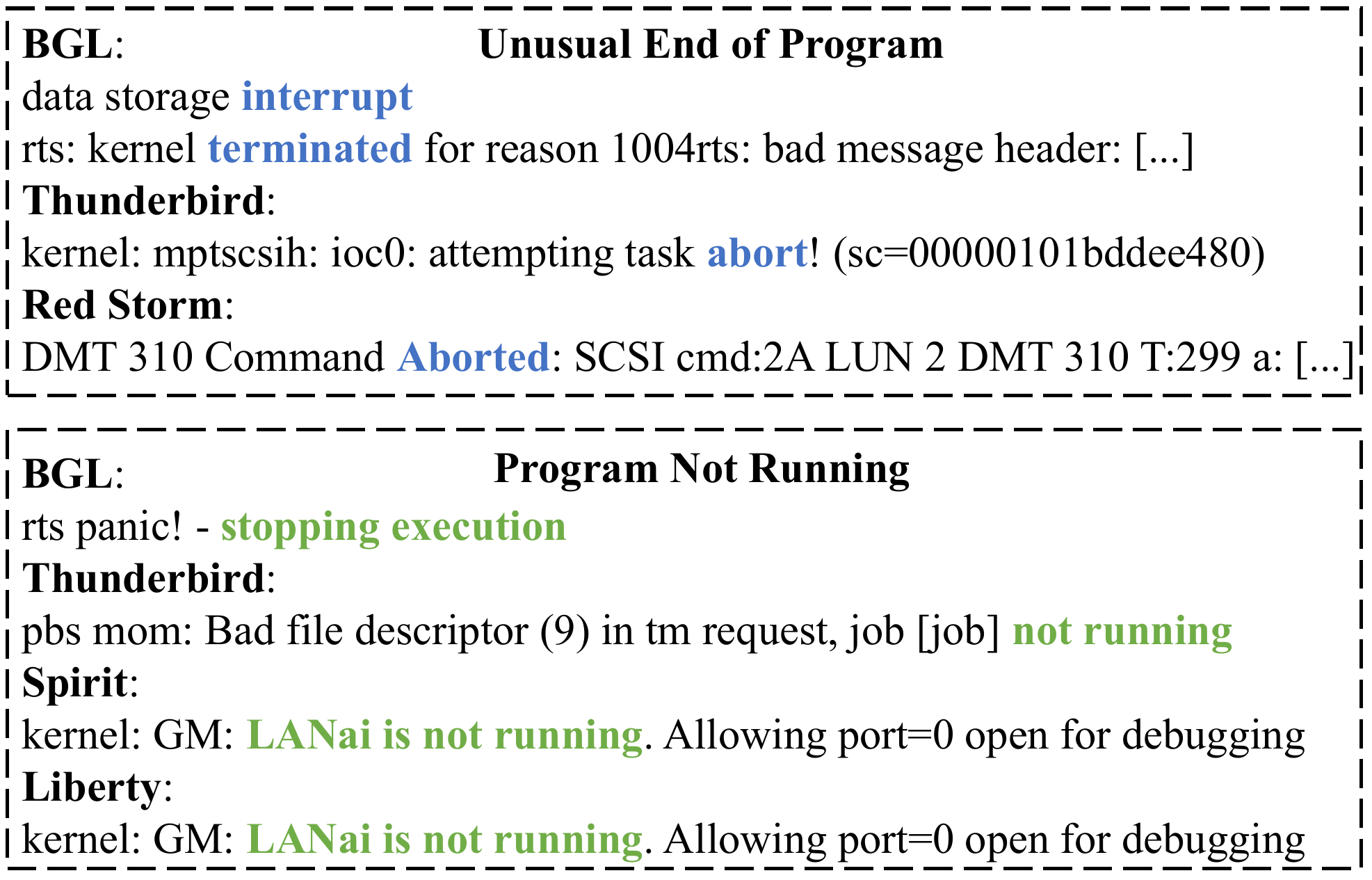}
    }
    \caption{The same anomaly from multiple domains. The top part denotes the ``Unusual End of Program'' anomaly from three domains including BGL, Thunderbird, and Red Storm while the bottom part is the ``Program Not Running'' from four domains including BGL, Thunderbird, Spirit, and Liberty.}
    \label{anomalyindomain}
    \vspace{-10pt}
\end{figure}

Large-scale services are usually implemented by hundreds of developers, it is error-prone to detect anomalous logs from a local perspective. In this case, some automatic detection methods based on machine learning are proposed \cite{xu2009detecting}. Due to the development of IT services, the volume of log data has grown to the point where traditional approaches are infeasible. Therefore, research has turned to deep learning methods \cite{zhang2016automated,du2017deeplog,zhang2019robust,meng2019loganomaly}. As log messages are half-structured and have their semantics, which is similar to natural language corpus, language models like LSTM and Transformer are leveraged to obtain semantics in logs. Recently proposed methods even adopt fashion pretrained models like BERT \cite{devlin2018bert}, GPT2 \cite{gpt2} for better embedding representation.

We observe that logs from different sources have the same anomalous categories. Despite being different in morphology and syntax, logs of multiple domains are semantically similar.
For example, in Figure \ref{anomalyindomain}, three sources (BGL, Thunderbird, Red Storm) all have the anomaly called the unusual end of program, thus we naturally think if the model can identify the same anomalies in all domains with shared semantic knowledge. However, existing approaches mostly focus on a single domain, when new components from a different/similar domain are introduced to the system, they lack the ability to accommodate such unseen log messages. In addition, we need to consider the continuous iteration of log data when system upgrades, which is costly to retrain different copies of the model. Therefore, a method based on transfer learning is required to perform well on logs from multiple domains.

In this paper, we address the problems above via a two-stage solution called \ourmethod{}. \ourmethod{} is capable of preserving the shared semantic knowledge between different domains. More specifically, we first create a neural network model based on the self-attention mechanism, which is pretrained on the source domain to obtain common semantics of log sequences. Second, \ourmethod{} utilizes a flexible plugin-in component called adapter to transfer knowledge from the source domain to the target domain. 

Generally, the main contributions of this work are listed as follows: (i) We propose \ourmethod{}, an end-to-end framework using Transformer encoder architecture to automatically detect log anomalies. (ii) With only a few additional trainable parameters on the target domain, \ourmethod{} allows a high degree of parameter-sharing while reducing time and calculation consumption. (iii) Our \ourmethod{} performs well under different amounts of training data, especially when the training data is low-resource. (iv) The proposed approach is evaluated against three public datasets: HDFS, BGL, and Thunderbird. \ourmethod{} reaches state-of-the-art performance on all these three datasets.
%%%%%%%%%%%%%%%%%%%%%%%%%%%%%%%%%%%%%%%%%%%%%%%%%%%%%%%%%%%%%%%%

% The purpose of log parsing is converting unstructured log data into the structured event template by removing parameters and keeping the keywords. 
% For example, in Figure\ref{preprocessing}, the log template “section closed for user $<*>$” can be extracted from the first log message. Here, “$*$” denotes the position of a parameter, which represents a specific username. After extracting all the templates, we match each log message and the corresponding template. Next, the log template sequence is fed into anomaly detection models.

% The previous work \cite{2017drain} adopts a fixed-depth tree structure to split the log data and extracts structured templates. AEL \cite{AEL} clusters log messages by comparing the frequency of occurrence between constants and variables. Spell \cite{spell2016} applies the longest common subsequence algorithm to parse logs efficiently. IPLoM \cite{Iplom} utilizes an iterative partition strategy to divide log messages into groups based on message length, token position, and mapping relationship.

\section{Background}
\begin{figure}[ht]
    \centering {
        \includegraphics[width=1\columnwidth]{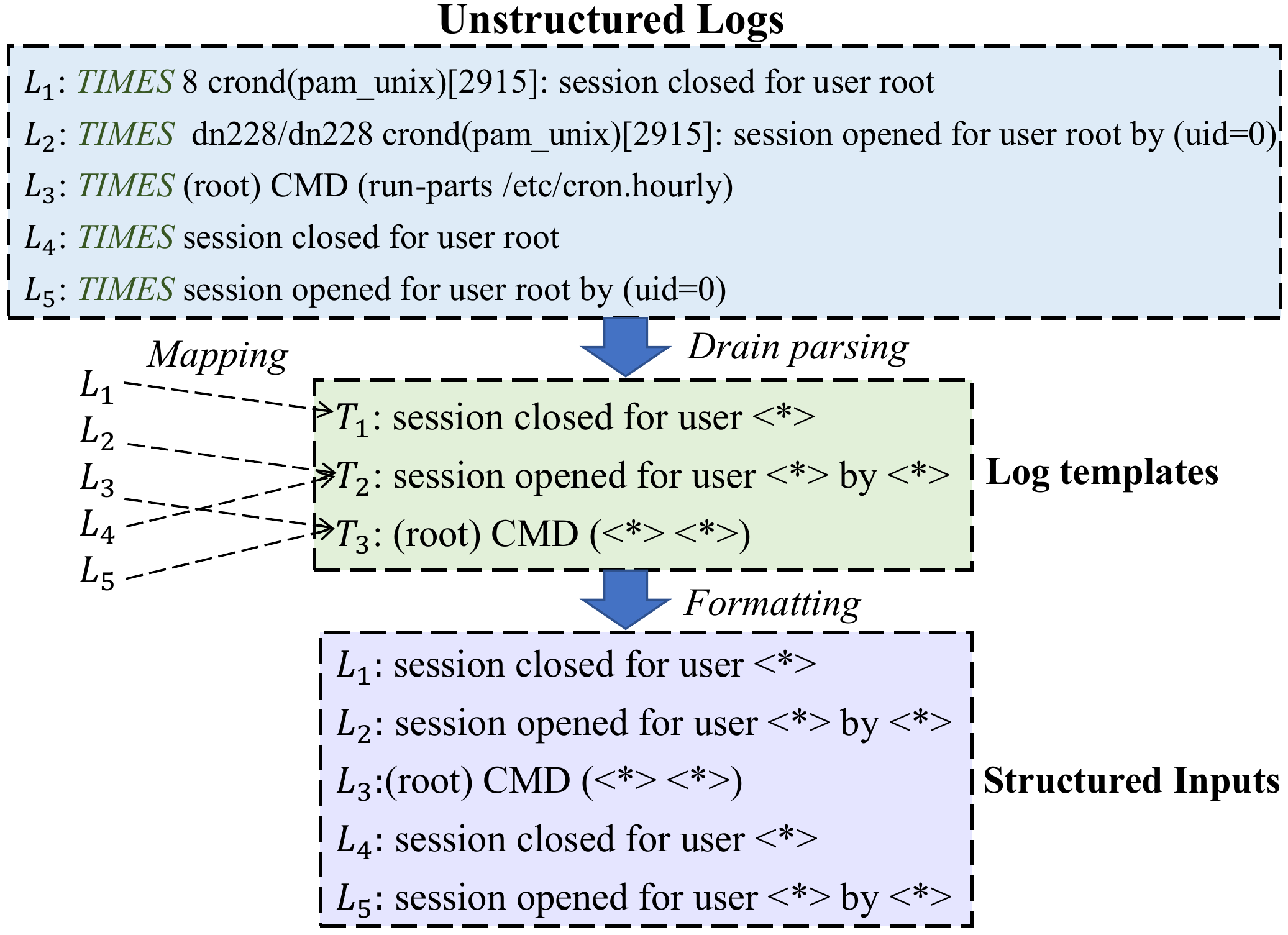}
    }
    \caption{Logs and Templates. The top part is unstructured logs, we adopt Drain algorithm to extract log templates,then we match each log with its template, which is the middle part. The bottom part is structured inputs.  }
    \label{preprocessing}
\end{figure}

\paragraph{Log Parsing}
The purpose of log parsing is to convert unstructured log data into the structured event template by removing parameters and keeping keywords \cite{AEL,Iplom}. The previous work Drain \cite{2017drain} adopts a fixed-depth tree structure to split the log data and extracts structured templates. Spell \cite{spell2016} applies the longest common subsequence algorithm to parse logs efficiently. In Figure \ref{preprocessing}, we utilize Drain to extract all the templates, and then each log message and the corresponding template is matched. Next, the whole log template sequence is fed into anomaly detection models.

\paragraph{Adapter-based Tuning}
Adapter-based tuning \cite{houlsby2019parameter,simpleadaption2019} is proved to be a parameter-efficient alternative in many NLP tasks. The structure of adapters is lightweight, which is usually composed of simple projection layers. Adapters are always inserted between transformer layers \cite{vaswani2017attention}. When tuning the model on downstream tasks, only the parameters of adapters are updated while the weights of the pretrained model are frozen. Thus, though utilizing much less trainable parameters compared to full fine-tuning \cite{devlin2018bert,Stickland019multitask}, adapter-based tuning reaches comparable performance. In this paper, we design our adapter layer as one down-projection layer, one activation layer, and one up-projection layer in Figure \ref{adapter}.

\paragraph{Log Anomaly Detection}

There are two main methods for log anomaly detection, including supervised and unsupervised methods. Supervised methods are often classification-based methods \cite{breier2015anomaly,hitanomaly,lu2018detecting,LogLAB}. LogRobust \cite{zhang2019robust} utilizes both normal and abnormal log data for training based on the Bi-LSTM architecture. Furthermore, Neurallog \cite{nurallog} uses BERT to transform raw log messages into semantic embeddings without log parsing. However, obtaining system-specific labeled samples is costly and impractical. Some unsupervised methods \cite{xu2009detecting,semi-supervisedloganomaly,A2log} have been proposed to alleviate such burden. DeepLog \cite{du2017deeplog} utilizes the LSTM network to forecast the next log sequence with the ranked probabilities. Besides, LogAnomaly \cite{meng2019loganomaly} utilizes the embeddings of logs to capture the semantic information. 

Although these methods attain the improvement of performance, they ignore sharing semantics between multiple log sources, mainly focusing on tackling the single log source setting. Our \ourmethod{} leverages such semantic knowledge efficiently based on the Transformer-adapter architecture.

\section{\ourmethod{}}

\begin{figure*}[t]
    \centering {\includegraphics[width=0.68\textwidth]{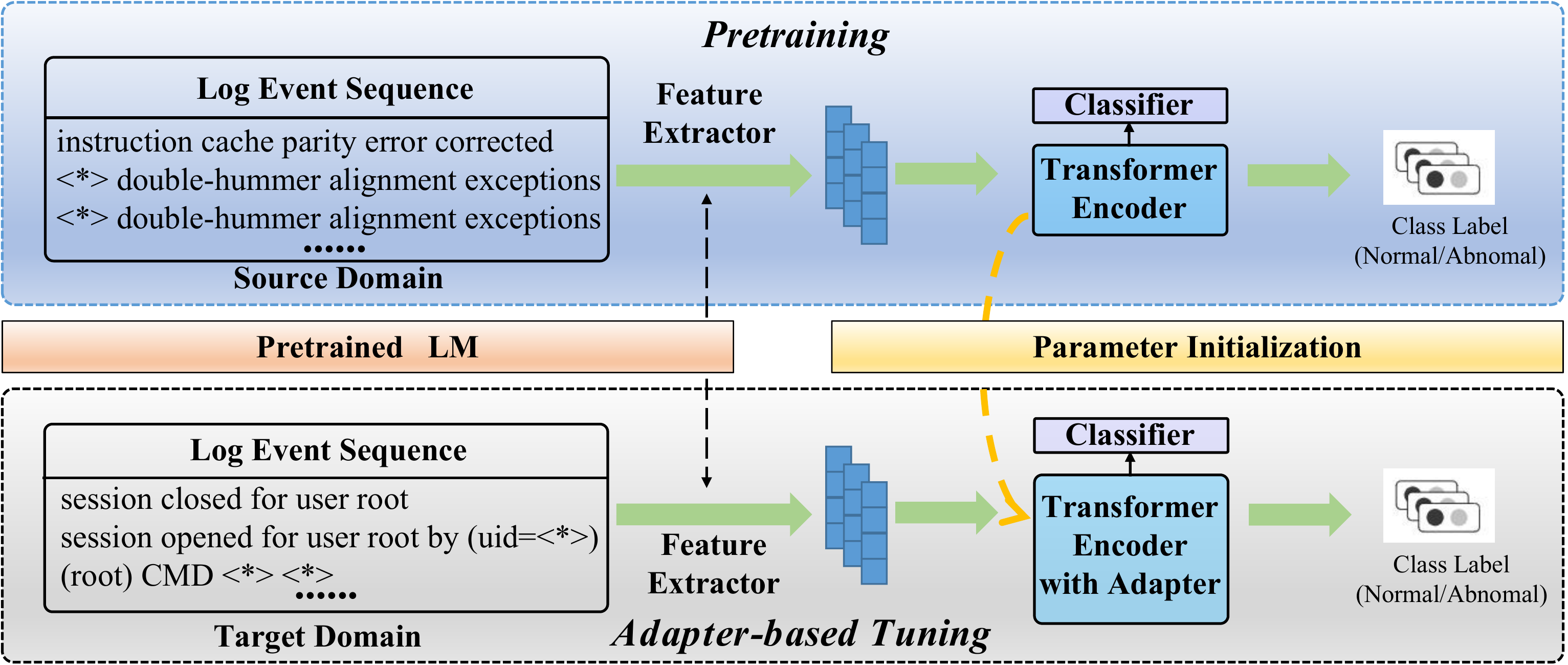}}
    \caption{Overview of our proposed architecture. All log event sequence is first fed into the pretrained language model to extract the representations. The Transformer encoder is trained on the high-resource source-domain dataset to acquire shared semantic information. Then, we initialize the Transformer encoder and only tune the parameters of the adapter on the target-domain dataset to transfer the knowledge from the source domain to the target domain.}
    \label{fig1}
\end{figure*}

In this section, we describe the general framework for log anomaly detection, named \ourmethod{}. The architecture of the \ourmethod{} is shown in Figure \ref{fig1}, which contains two stages: pretraining and adapter-based tuning. The following parts start with the definition of the problem, and then the components of the backbone model are presented. Afterward, we illustrate the exhaustive procedure of two stages.

\subsection{Problem Definition}

Log anomaly detection problem is defined as a dichotomy problem. The model is supposed to determine whether the input log is abnormal or normal. For the source domain, assuming that through preprocessing, we achieve the vector representations of $K_{src}$ log sequences, which is denoted as $S^{src}=\{ S_{k}\}_{k=1}^{K_{src}}$. Then, $S^{src}_{i}=\{V_{t}^{src}\}_{t=1}^{T_i^{src}}$ denotes the $i$-th log sequence, where $T_i^{src}$ is the length of the $i$-th log sequence.
For the target domain, $S^{tgt}= \{S^{tgt}_{k}\}_{k=1}^{K_{tgt}}$ denotes the representations of $K_{tgt}$ log sequences. $S^{tgt}_{j}=\{V_{t}^{tgt}\}_{t=1}^{T_i^{tgt}}$ denotes the $i$-th log sequence, where $T_i^{tgt}$ is the length of the $i$-th log sequence.
Therefore, the training procedure is defined as follows. We first pretrain the model on the source-domain dataset as below:
\begin{equation}
    f_{p}(y_{i}|S^{src}_{i};\Theta)), 
    \label{eq:1}
\end{equation}where $f_{p}$ represents the pretraining stage,$\Theta$ is the parameter of the model in pretraining stage. Then, the model is transferred to the target-domain as below:
\begin{equation}
    f_{a}(y_{j}|S^{src}_{j};\Theta_{f},\theta_{a}). 
    \label{eq:2}
\end{equation}where $f_{a}$ represents the adapter-based tuning stage. $\Theta_{f}$ is the parameter of the transformer encoder transferred from the pretraining stage, which is frozen in adapter-based tuning stage. $\theta_{a}$ is the parameter of the adapter. $y$ is the groundtruth. Through Equation \ref{eq:1} and \ref{eq:2}, \ourmethod{} learns the semantic representation of template sequences between domains.

%\subsection{\ourmethod{} Architecture}
\subsection{Backbone Model}
%
% adapter target domain 
\paragraph{Feature Extractor} %BERT

% \subsubsection{Embedding Layer}
The feature extractor converts session sequences (template sequence) to vectors with the same dimension $d$. Here we use the pretrained sentence-bert\cite{reimers2019sentence} model to get the template sequence representation. Recently some methods extract semantic representation from raw log messages, they believe it could prevent the loss of information due to log parsing errors. However, embedding every log message is not realistic considering a large amount of log data. Studies also show that almost all anomalies could be detected by template sequence, even if there are parsing errors. Thus, we only embed all existing log templates. Each session has $l$ fixed length, so through the layer, we can obtain the $X \epsilon R_{}^{l \times d}$ for each session. 

\paragraph{Encoder with Light Adapter}
\begin{figure}[ht]
    \centering {
        \includegraphics[width=0.5\columnwidth]{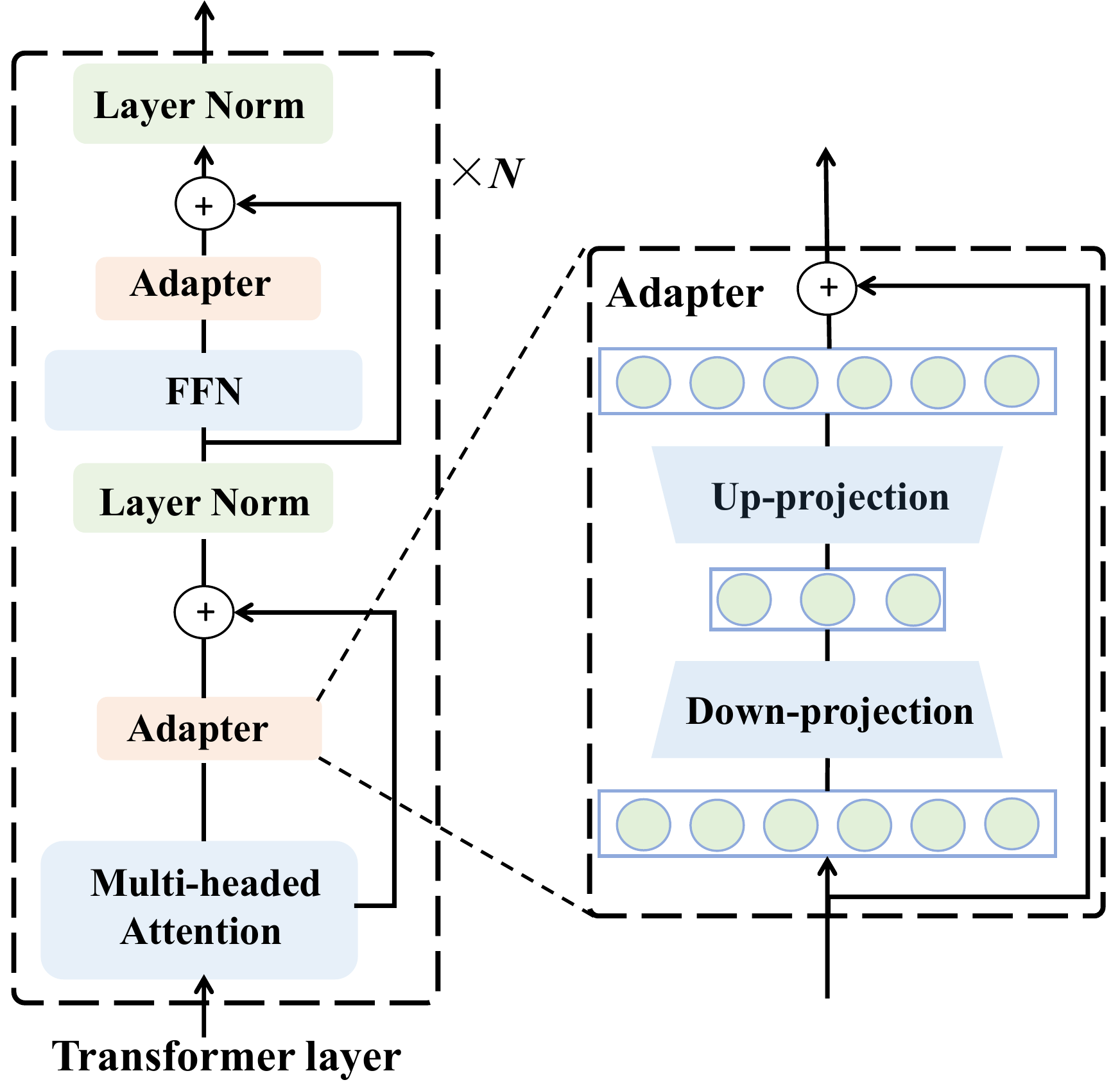}
    }
    \caption{Encoder with light adapter. Where $N$ is the number of transformer layers. The left part describes the traditional transformer encoder inserted by adapters, the right part is our light adapter, which is composed of the down- and up-projection layers.}
    \label{adapter}
\end{figure}

As shown in Figure \ref{adapter}, To better encode the corresponding feature of inputs, we use the transformer encoder as the backbone model. By doing so, our encoder with self-attention mechanism overcomes the limitations of RNN-based models. The core self-attention mechanism is formally written as:
\begin{equation}
    Attention(Q,K,V)=softmax(\frac{QK_{}^{T}}{\sqrt{d/heads} } )V.
\end{equation}
where $heads$ is the number of the heads, $d$ denotes the dimension of the input, and $Q,K,V$ represent queries, keys, and values, respectively. 

The order of a log sequence conveys information of the program execution sequence. Wrong execution order is also considered abnormal. Thus, constant positional embedding is also used.Component after self-attention layer and feedforward layer is the adapter. It's a lightweight neural network between the transformer layers. When tuning a pretrained language model to a new domain, adapters are inserted. During adapter-based tuning, only a few parameters of the adapters are updated on the target domain. More specifically, we use down- and up-scale neural networks as the adapter. Two projection layers in adapter first map hidden vector from dimension $d$ to dimension $m$ and then map it back to $d$. The adapter also has a skip-connection operation internally. The output vector $h'$ of the adapter is calculated as follow:
\begin{equation}
    h'= W_{up} tanh(W_{down}h)+h.
\end{equation}
where $h \in \mathbb{R}^d$ represents a given hidden vector.  $W_{down} \in \mathbb{R}^{m\times d}$ and $W_{up} \in \mathbb{R}^{d\times m}$ is the down-projection and the up-projection matrix respectively, by setting $m<<d$, we limit the number of parameters added per adapter, which is the core to reduce trainable parameters while retaining semantic information to the maximum extent.

\subsection{Pretraining}
Inspired by the BERT model \cite{devlin2018bert}, which takes a self-supervised method to learn general language features that further be utilized to serve different downstream tasks, we acquire the common reason for log anomaly with the stacked transformer encoder. In this stage, the pretrained model learns the commonalities among different anomalies from the semantic level, which contributes significantly to the anomalous detection for new log sources.
More specifically, the objective of this stage is the same as anomaly detection, which is a supervised classification task without adapters in the model. Then, the parameters of the transformer encoder, which is trained during this stage, are shared to the next stage. After parameter initialization, we freeze these parameters during the adapter-based tuning stage.

\subsection{Adapter-based Tuning}
\label{adapter_based_tuning}
When tuning a pretrained model from the source domain to a target domain, the way of adapter-based tuning leverages the knowledge obtained from the pretraining stage with lightweight adapters, which are neural networks like \cite{houlsby2019parameter}. 
In this paper, our adapter is composed of one down-projection layer, one activation layer, and one up-projection layer in Figure \ref{adapter}.
Through the pretraining stage, we achieve the pretrained model, thus in this second stage, we plug adapters into the transformer layers of the pretrained model, afterward, only the parameters of the adapters are updated during target domain adaption. Parameters of the multi-headed attention and the feedforward layers in the pretrained model are frozen. Unlike fine-tuning, \ourmethod{} provides a plug-in mechanism to reuse the pretrained model with only a few additional trainable parameters, without updating the entire model for a new domain in this stage.  

%Section \ref{adapter_based_tuning}
% By adding adapter to the model, the probability of prediction becomes:
% $$p(x|\theta_a;\theta)=f(x;\Theta,\theta_a)$$
% where $\theta$ is the parameters of transformers excluding the final classifier, and $\theta_a$ is the parameters of adapters and final classifier, which is obtained by maximizing the production:

% $$\theta_a=argmax_{\theta_a}\prod f(x;\theta,\theta_a)$$

\subsection{Training Strategy}
In this work, we both adopt BCE loss for two stages. Thus, we define the objective loss of the pretraining stage as below.
\begin{equation}
       \mathcal{L}_{p} = -\mathbb{E}_{x,y \in D^{src}_{x,y}} [\log P(y|x;\Theta) ], 
\end{equation}where $\mathcal{L}_{p}$ represents the loss in the pre-training stage. $\Theta$ is the parameter of the whole model in the pretraining stage. $x$ and $y$ are the input data and label respectively, $D^{src}_{x,y}$ represents the data coming from the source domain. 
Then, we define the objective loss in the adapter-based tuning stage as below. 
\begin{equation}
       \mathcal{L}_{a} = -\mathbb{E}_{x,y \in D^{tgt}_{x,y}} [\log P(y|x;\Theta_{f},\theta_{a}) ]. 
\end{equation}where $\mathcal{L}_{a}$ is the loss function in the adapter-based tuning stage. $\Theta_{f}$ is the parameter of the encoder module trained in the pretraining stage, which is frozen in the adapter-based tuning stage. $\theta_{a}$ is the parameter of the adapter. $D^{tgt}_{x,y}$ represents the data coming from the target domain.

\section{Experiments}
%This section describes the datasets and evaluation metrics employed
%for our experiments, followed by introduction of baseline models
%and implementation details. Performances of our \ourmethod{} and an ablation study of the model structure is then presented.

% In this section, we first experiment with three public datasets from loghub\cite{he2020loghub}, namely HDFS\cite{xu2009detecting}, Blue Gene/L and Thunderbird\cite{oliner2007supercomputers}, which shows that our purposed \ourmethod{} model achieves the state-of-the-art performance in traditional log anomaly detection. 
% Then we find that fine-tuning using transferred parameters, model converges much faster than training from scratch and is more stable. It confirms the the feasibility of transfer learning between different log data sources. 
% Finally, we show that using only $3.5\%-5.5\%$ trainable parameters, \ourmethod{} with adapter\cite{houlsby2019parameter} achieves almost the same performance as full fine-tuning.

In this section, the comprehensive settings of the experiment are illustrated. Afterward, we experiment on three public datasets coming from LogHub \cite{he2020loghub}. Compared with baseline methods, our \ourmethod{} reaches the state-of-the-art performance on all datasets.

% which confirms the feasibility of transfer learning between different log data sources. 

% \begin{table*}[ht]
% % 下面一行是字体大小
%     \normalsize
%     % 列与列之间的距离
%     \setlength{\tabcolsep}{2mm}
%     \begin{center}
%         % 设置每一列的对其方式，l是左对齐，c是居中
%         \begin{tabular}{lcccc}
%             \hline
%             Name   & Category   &   \#Messages     & \#Anomaly samples & \#Templates   
%             \\
%             \hline
%             HDFS   &  Distributed system   &    11,175,629  &  16,838 & 49     
%             \\
%             BGL    &    Supercomputer     & 4,747,963       & - & 423    
%             \\
%             Thunderbird   & Supercomputer  &  10,000,000    & - & 1292
%             \\
%             \hline
%         \end{tabular}
%         \caption{A summary of the datasets used in this work. Messages are the raw log strings. Samples are log sequences extracted by ID or sliding window of size 20. In the following experiment, we consider BGL and Thunderbird as same domain since they are both supercomputer logs.}
%         \label{table1}
%     \end{center}
% \end{table*}

\begin{table}[t]
    \begin{center}
        % 设置每一列的对其方式，l是左对齐，c是居中
        \resizebox{0.9\columnwidth}{!}{
        \begin{tabular}{lcccc}
            \toprule
            Dataset   & Category   &   \#Messages     & \#Anomaly & \#Templates   
            \\
            \midrule
            HDFS   &  Distributed   &    11M  &  17K & 49     
            \\
            BGL    &    Supercomputer     & 5M       & 20K   & 423    
            \\
            Thunderbird   & Supercomputer  &  10M    & 123K  & 1292
            \\
            \bottomrule
        \end{tabular}}
        \caption{A summary of the datasets used in this work. Messages are the raw log strings. Samples are log sequences extracted by ID or sliding window of size 20.}
        \label{table1}
    \end{center}
\end{table}

\paragraph{Datasets} We conduct experiments on three public datasets, which is described in Table \ref{table1}. 10M/11M/5M continuous log messages from Thunderbird/HDFS/BGL are separately leveraged, which is used in prior work \cite{yao2020study,nurallog}. HDFS \cite{xu2009detecting} dataset is generated and collected from the Amazon EC2 platform through running Hadoop-based map-reduce jobs. It contains messages about blocks that assign a unique ID to the raw logs.
Thunderbird and BGL datasets\cite{oliner2007supercomputers} contain logs collected from a two supercomputer system at Sandia National Labs (SNL) in Albuquerque. The log contains alert and non-alert messages identified by alert category tags. Each log message in the datasets was manually labeled as anomalous or not.

\paragraph{Preprocessing}

Different datasets require preprocessing correspondingly. We extract log sequences by block IDs for HDFS, since logs in HDFS with the same block ID are correlated. BGL and Thunderbird do not have such IDs, so we utilize a sliding window(size of 20) without overlap to generate a log sequence. \ref{table1} shows the detail of datasets. We adopt Drain\cite{2017drain} with specifically designed regex to do log parsing, due to its high efficiency. Number of anomaly is counted based on window. Windows containing anomalous message are considered as anomalies, thus it's less than the number of anomalous log messages. For each dataset, we select the first $80\%$ (according to the timestamps of logs) log sequences for training and the rest $20\%$ for testing.

\paragraph{Implementation Details}
In the experiment, we try a different number of transformer encoder layers in $\{1,2,4\}$. The number of attention heads is 8, and the size of the feedforward network that takes the output of the multi-head self-attention mechanism is 3072. 
We optimize using Adam way whose learning rate is scheduled by OneCycleLR, with $\beta_1 = 0.9$, $\beta_2 = 0.99$, and $\varepsilon = 10^{-8}$. All runs are trained on 4 NVIDIA v100 with a batch size of 64. For each dataset, we tune the maximum learning of OneCycleLR scheduler in $\{1e-5,5e-5,1e-6\}$.

%These evaluation metrics can reflect the performance of the considered methods from different aspects. 

% Precision is used to compute the percentages of how many reported anomalies are correct. Recall shows how many anomalies are detected among the true anomalies set. While $F_1$ score is harmonic mean value of Precision and Recall. 

\paragraph{Baselines and Evaluation}
We compare \ourmethod{} with the six baseline methods, including Logistic Regression(LR), Support Vector Machine(SVM), Deeplog \cite{du2017deeplog}, LogAnomaly \cite{meng2019loganomaly}, LogRobust \cite{zhang2019robust} and Neurallog \cite{nurallog} on the three datasets. These methods are in two categories: machine learning and neural network approaches. Traditional approaches usually build models by transforming the log sequence into log count vectors while neural network approaches leverage word or contextual embeddings to represent log sequences. In our experiments, we use precision ($\frac{TP}{TP+FP}$), recall ($\frac{TP}{TP+FN}$) and $F_1$ score ($\frac{2*Precision*Recall}{Precision+Recall}$) to compare our method and previous baselines.

\paragraph{Main Results}

\begin{table}[t]
    \centering
    \begin{center}
    \resizebox{0.8\columnwidth}{!}{
    \begin{tabular}{l c c c c}
        \toprule
        Dataset & Method & Precision & Recall & $F_1$ Score\\
        
        \midrule

        &LR  & 0.96 & 0.91  &0.93 \\
        &SVM  &0.96  &0.97  & 0.97 \\
        &DeepLog  & 0.95  & 0.96  & 0.96 \\
        HDFS&LogAnomaly & 0.96  & 0.94 & 0.96   \\
        &LogRobust  & 0.98  & 0.98  & 0.98 \\
        &Neurallog  & 0.96  & \textbf{1}  & 0.98 \\
        &\ourmethod{}  & \textbf{0.99}  & 0.99  & \textbf{0.99} \\
        % &Translog$_{BGL}$  & 0.9966  & 0.9989  & \textbf{0.9977} \\
        % &Translog$_{Thunderbird}$  & 0.9957  & 0.9980  & 0.9969 \\
            
        \midrule
        
        & LR & 0.78 &0.79 &0.78\\
        &SVM &0.89 &0.86&0.87\\
        &DeepLog& 0.90 & 0.83 &0.86\\
        BGL&LogAnomaly & 0.97 &0.94&0.96\\
        &LogRobust& 0.62 & 0.96 &0.73\\
        &NeuralLog & 0.61 & 0.78 &0.68\\
        &\ourmethod{} & \textbf{0.98} & \textbf{0.98} & \textbf{0.98}\\
        % &Translog$_{Thunderbird}$ & \textbf{0.9914} & 0.9161 & 0.9522\\
            
        \midrule
        
        &LR & 0.46 & 0.91 &0.61\\
        &SVM & 0.34 & 0.91 &0.50\\
        &DeepLog & - & - & -\\
        Thunderbird &LogAnomaly & 0.61 & 0.78 &0.68\\
        &Logrobust & 0.61 & 0.78 &0.68\\
        &NeuralLog & 0.93 & \textbf{1} &0.96\\
        &\ourmethod{} & \textbf{0.99} & 0.99 & \textbf{0.99}\\
        % &Translog$_{BGL}$ & 0.9948 & 0.9978 & 0.9963\\
        \bottomrule
    \end{tabular}}
    \caption{Experimental results compared with baseline models on Thunderbird, BGL and HDFS. The best results are highlighted.}
    \label{table2}
    \end{center}
\end{table}

To test the effectiveness of \ourmethod{}, we compare the proposed algorithm with baseline methods on HDFS, Thunderbird, and BGL benchmarks. As is shown in TABLE. \ref{table2}.  Our \ourmethod{} achieves the highest $F_1$ score on all three datasets, confirming the effectiveness and generalization of  \ourmethod{}. To obtain our main results, BGL is pretrained as the source domain in the first stage, while HDFS and Thunderbird are selected as the target domain in the adapter-based tuning stage. We provide a thorough analysis of our model in Section \ref{analysis}.

% In this section, we study the influence of using different ratios of the training data. 
% We choose four input data ratios during our experiment: 10$\%$, 50$\%$, 70$\%$, and 100$\%$. 
% The experimental results of the three datasets are shown in Fig.\ref{fig3}, Fig.\ref{fig4}, and Fig.\ref{fig5}. Experiments show that our model has the under-fitting problem when the amount of input data is too small for our model to learn enough knowledge, with the increase of the input data, the result becomes better.
% % 添加的部分
% It is worth noting that in our ablation experiments, this trend does not seem to have reached saturation
% Based on this, we conclude that our model has the characteristic called data hungry, which makes our model more powerful in real-world industry scenarios.
\section{Analysis} \label{analysis}
In this section, we conduct the ablation study in four aspects for a penetrating analysis of \ourmethod{}, including the effect of the pretrained model, the gap between pretrained log models, the efficiency of adapter-based tuning, and the low-resource study.

% \begin{figure*}[ht]
%     \centering
%     \subfloat[HDFS]{\includegraphics[width=0.46\textwidth]{transfer_hdfs.png}\label{fig:f1}}
%     \hfill
%     \subfloat[Thunderbird]{\includegraphics[width=0.46\textwidth]{transfer_thunderbird.png}\label{fig:f2}}
%   
%     \caption{Loss and $F_1$ score during training w.r.t. the number of batches. Result of fine-tuning are based on 1-layer \ourmethod{} trained on BGL dataset. Both of training and fine-tuning use the same learning rate.}
%     \label{fig3}
% \end{figure*}

\begin{figure}[ht]
    \centering {
        \includegraphics[width=1\linewidth]{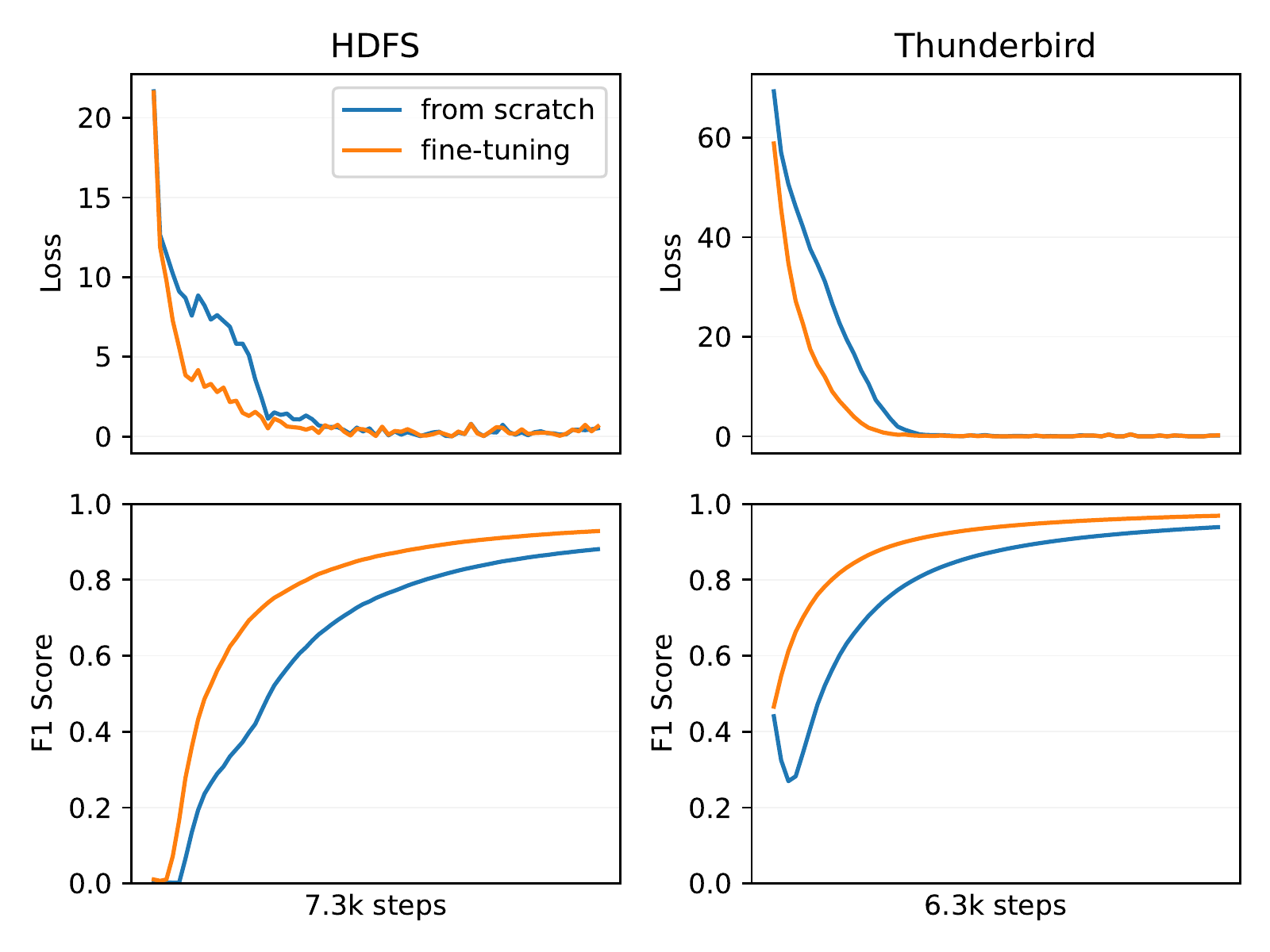}
    }
    \caption{Loss and $F_1$ score on the dev set w.r.t training steps. The left/right part represents Loss/$F_1$ score of the HDFS/Thunderbird. We compare two ways of training including training from scratch and fine-tuning from model pretrained on BGL. All results are using 1-layer Transformer encoder and the same learning rate.}
    \label{fig:transfer}
\end{figure}

% \begin{figure}[ht]
%     \centering {
%         \includegraphics[width=\linewidth]{pretrainedmodel.pdf}
%     }
%     \caption{Loss on the dev set w.r.t training steps. The upper/bottom results are based on parameters pretrained on BGL/Thunderbird, thus BGL/Thunderbird are not shown. All results are using 1-layer Transformer encoder and the same learning rate.}
%     \label{fig:pretrainedmodel}
% \end{figure}

\paragraph{Effect of Pretrained Log Model} \label{effectlogmodel}
To demonstrate the feasibility of transferring semantic information between various domains, we compare the performance of two training techniques, namely training from scratch and fine-tuning. Here, the way of fine-tuning is to update the parameter of the whole pretrained model. We choose BGL as the source domain for its variety in log templates and its huge data volume. Besides, from the perspective of system kinds, HDFS is a distributed system while Thunderbird is a supercomputer system similar to BGL. Thus, experiments on such different log data distributions can demonstrate both the transferability of semantic knowledge and the effectiveness of \ourmethod{}.

We compare two methods in terms of their rate of convergence and final results. Figure \ref{fig:transfer} displays the loss curves and $F_1$ score curves w.r.t training steps. The results show that fine-tuning converges faster than training from scratch, which demonstrates that semantic knowledge the pretrained model learned is valuable, besides it also shows the power of the model to transfer information between domains.  We furthermore discover that fine-tuning is more stable with a smoother curve, which is more obvious on Thunderbird, illustrating that similar domains share semantics more efficiently. 

Due to \ourmethod{} achieves good performance even trained from scratch, the final $F_1$ scores of two methods are close. By observing the $F_1$ score curve, we discover that fine-tuning requires fewer training steps to gain the best result, which is noteworthy for reducing costs in industrial scenes.

\paragraph{Gap between Pretrained Log Models} 
% 预训练模型本身的性能
\begin{figure}[ht]
    \centering {
        \includegraphics[width=\linewidth]{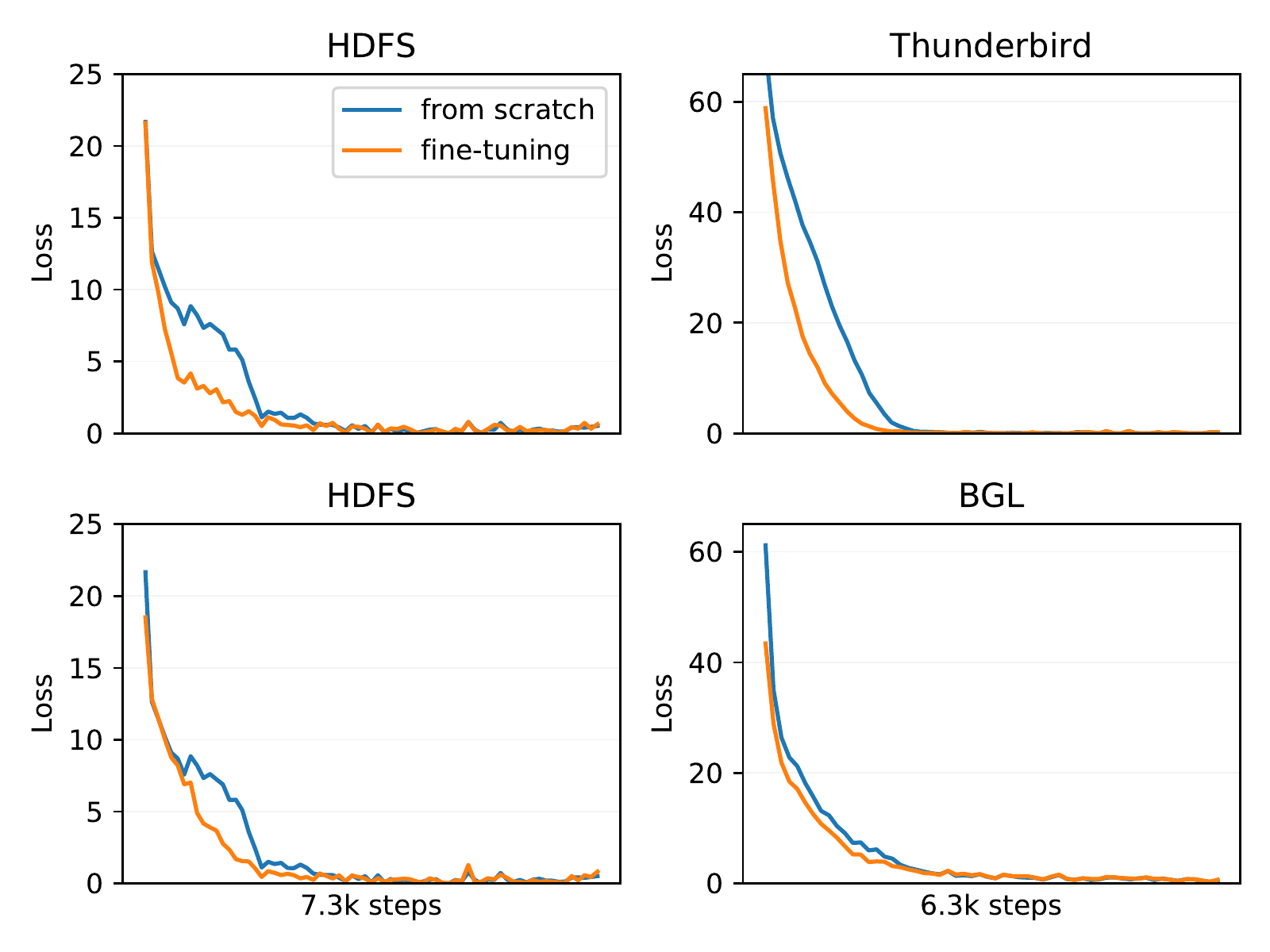}
    }
    \caption{Loss on the dev set w.r.t training steps. The upper/bottom results are based on parameters pretrained on BGL/Thunderbird, thus BGL/Thunderbird are not shown. All results are using 1-layer Transformer encoder and the same learning rate.}
    \label{fig:pretrainedmodel}
\end{figure}

We analyzed the effectiveness of the pretrained log model. The experimental results show the feasibility of pre-training, but we can not explain whether the performance improvement comes from the pretraining way or the pretrained model itself.  Therefore, in this part, we adopt different pretrained log models to analyze the gap between pretrained log models.
Specifically, we utilize BGL and Thunderbird as the source data respectively. Figure \ref{fig:pretrainedmodel} shows the loss curves of the two training strategies on the target source. From the curves, we can see that the fine-tuning loss of each model decreases faster than training from scratch. At the same time, comparing the two pre-training models, it is obvious that, for the HDFS dataset, the model pretrained on BGL brings greater performance improvement. In conclusion, different pre-training models provide different gains for model performance, which is the gap between pretrained log models.

\paragraph{Efficiency of Adapter-based Tuning}

Although we have verified that parameters transfer could accelerate convergence without reducing performance, fine-tuning each component is expensive and inconvenient. Thus, we adopt a parameter-efficient strategy, called adapter-based tuning, to allow a high degree of sharing knowledge between domains. By utilizing the adapter, we acquire a compact model for log anomaly detection by adding a few additional parameters. 

To confirm the efficiency of our \ourmethod{}, general transfer performance of fine-tuning and adapter-based tuning are compared. Experiments are based on a model trained on the BGL and Thunderbird datasets. On each dataset, we utilize batch size 64 and tune the model in learning rate selected from $\{1e-5,5e-5,1e-6\}$. Results show that $5e-5$ is the most satisfactory for HDFS, $1e-5$ is the best for Thunderbird. We conduct the ablation study by adjusting the number of encoder layers in $\{1,2,4\}$. Table \ref{tab:transfer} summarizes the results. We observe that our \ourmethod{} with adapters generates a competitive score but adopts $3.5\%-5.5\%$ of the parameters in the whole original model. In addition, experiments on the various number of transformer encoders are executed.  Results indicate that more encoder layers for fine-tuning do not always generate better results. Simultaneously, adapter-based tuning performs more robust when we stack more encoder layers.  

\begin{table}
\centering
\resizebox{0.8\columnwidth}{!}{
\begin{tabular}{lrrrr}
\toprule
Method          & Layers & Parameters & HDFS & Thunderbird \\
\midrule

       & 1  &   7.2M   & 0.998  &  0.997    \\
Fine-tuning                & 2  &   14.3M    & 0.998  &  0.997    \\
                & 4  &   28.5M    & 0.997  &  0.998    \\
                \midrule
             & 1  &   0.4M   &  0.997  &  0.990  \\
Adapter                & 2  &   0.6M    &  0.997  &  0.996    \\
                & 4  &   1M    &  0.998  &  0.996    \\
  
\bottomrule
\end{tabular}}

\caption{Transfer result using fine-tuning and adapter based fine-tuning. \emph{Layers} is the number of transformer encoder layers. \emph{Parameters} is the number of trainable parameters in the model.}
\label{tab:transfer}
\end{table}

% \paragraph{Performance on Low-resource Setting}
\paragraph{Low-resource Study}

%\begin{figure}[ht]
%    \centering
%    \subfloat[]{\includegraphics[width=0.5\columnwidth]{low_hdfs.png}\label{fig:f1}}
%    \hfill
%    \subfloat[]{\includegraphics[width=0.5\columnwidth]{low_thunderbird.png}\label{fig:f2}}
%    \caption{Test performance w.r.t.the number of training examples. (a) HDFS. 5k, 10k, 20k, 50k corresponding to the font $1.1\%$, $2.2\%$, $4.3\%$, $10\%$training data respectively. (b) Thunderbird. 5k, 10k, 20k, 50k corresponding to font $2.5\%$, $5.3\%$, $10.6\%$, $27.4\%$training data. We sample these randomly, since the anomalies in Thunderbird mainly concentrate in the latter part of training data. }
%    \label{fig:low}
%    % \vspace{-5pt}
%\end{figure}

% \begin{figure}[ht]
%     \centering {
%         \includegraphics[width=0.75\linewidth]{low.pdf}
%     }
%     \caption{Test performance on BGL (pretrained on Thunderbird) w.r.t.the number of training examples. 5k, 10k, 20k, 50k corresponding to the font $2.5\%$, $5\%$, $10\%$, $25\%$ training data respectively. We show mean and standard deviation across 3 runs for all methods.}
%     \label{fig:low}
% \end{figure}

\begin{figure}[ht]
    \centering {
        \includegraphics[width=0.75\linewidth]{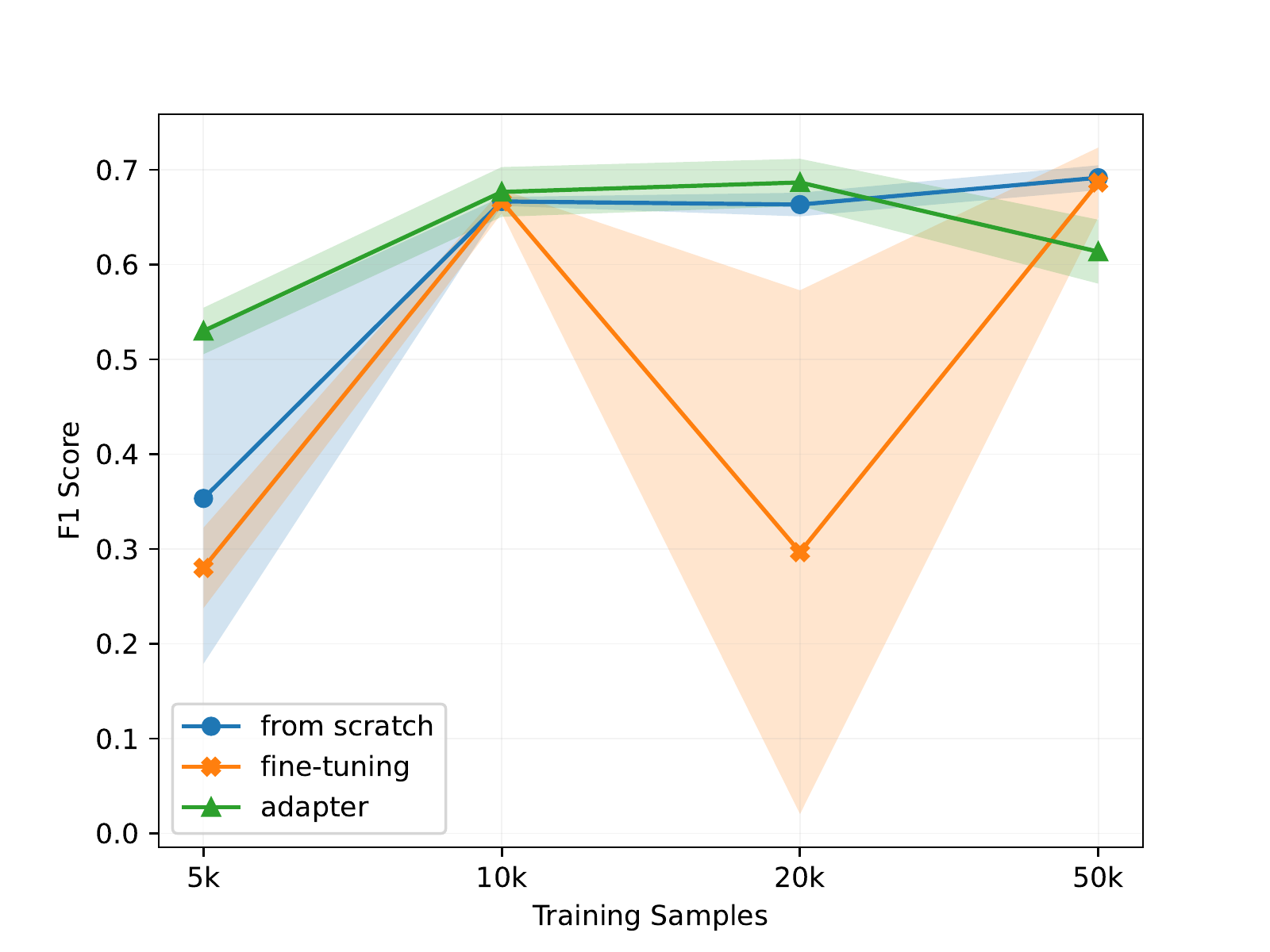}
    }
    \caption{Test performance on BGL (pretrained on Thunderbird) w.r.t.the number of training examples. 5k, 10k, 20k, 50k corresponding to the font $2.5\%$, $5\%$, $10\%$, $25\%$ training data respectively. We show mean and standard deviation across 3 runs for all methods.}
    \label{fig:low}
\end{figure}

To verify the influence of training size, we plot the performances with a varying number of training samples in Figure \ref{fig:low}. It presents the comparison results on the BGL dataset. We consider tasks with fewer than 50k training examples as low-resource tasks, as log data is easy to obtain and be labeled manually. We train models for 30 epochs to make sure they are sufficiently trained.

We find that adapter-based tuning consistently outperforms training and fine-tuning. The improvement is more significant when the training size is small. With the number of training samples increasing, both methods will gradually catch up and finally achieve similar results.
Another finding is that the quality of the model across runs is more robust, with a similar standard deviation across different training sizes. However, training and fine-tuning yield large variances when the number of the training sample is 5k and 20k.

To summarize, adapter tuning is highly parameter-efficient, and shared parameters contain semantic information that helps the model to detect anomalies. Training adapters with sizes less than $5\%$, performance decreases nearly $1\%$.

\section{Conclusion}
%Logs are one of the most valuable data in the scene of system operation and maintenance. Existing log anomaly detection methods fail to generalize well on unseen new log samples. 

In this paper, we propose \ourmethod{}, a unified transformer-based framework for log anomaly detection, which contains the pretraining stage and the adapter-based tuning stage. Extensive experiments demonstrate that our \ourmethod{}, with fewer trainable parameters and lower training costs, outperforms all previous baselines. We foresee the semantic migration between log sources for a unified multiple sources detection.

\clearpage
\bibliographystyle{named}
\bibliography{ijcai22}

\end{document}